\documentclass{bmvc2k}

\title{Geometry-Constrained Dynamic Hypergraph Convolutional Network with Contrastive Score Refinement for Skeleton-Based Action Recognition}

\addauthor{Minh Ta Tran Nhat}{nhatminhtatran@chungbuk.ac.kr}{1}
\addauthor{Yuning Ye}{yeyuning@chungbuk.ac.kr}{1}
\addauthor{Yuseok Ban}{ban@cbnu.ac.kr}{1}
\addinstitution{
 Department of Electronics Engineering \\
 Chungbuk National University\\
 Cheongju, South Korea
}

\runninghead{Ta Tran, Ye, Ban}{DyGCN}

\usepackage{multirow}
\usepackage[table]{xcolor}
\usepackage{tikz}
\usepackage{enumitem} 
\usetikzlibrary{arrows.meta, positioning}
\usepackage{amsmath}
\usepackage{amssymb}
 \usepackage{booktabs}
 \usepackage{bbm}
 \usepackage{svg}
 \usepackage{soul}
\begin{document}

\maketitle

\begin{abstract}
Skeleton-based action recognition is a pivotal task in computer vision; however, conventional Graph Convolutional Networks (GCNs) are often constrained by pairwise joint modeling, which fails to capture the intricate high-order interactions inherent in human movement. While recent hypergraph-based methods attempt to address this limitation, they typically rely on static structures derived from anatomical priors, lacking the flexibility to capture dynamic, action-specific joint correlations. To this end, we propose DyGCN, a Geometry-Constrained Dynamic Hypergraph Convolutional Network for skeleton-based action recognition. At the core of DyGCN is the Dynamic Hypergraph Construction Network with Contrastive Score Refinement (DyGCT) block, which comprises two novel components: the Geometry-Constrained Dynamic Hypergraph Construction (GeoHC) module and the Contrastive Score Refinement (CSR) module. Specifically, GeoHC dynamically constructs action-specific hyperedges by leveraging geometry-constrained second-order compatibility, transcending the limitations of static hyperedge structures. Furthermore, CSR explicitly refines the pairwise compatibility scores during hypergraph construction through prototype-based contrastive learning, fostering more discriminative action-specific hyperedges — with zero inference-time overhead as CSR operates only during training. Extensive experiments on NTU RGB+D, NTU RGB+D 120, and Northwestern-UCLA validate the effectiveness of DyGCN, achieving state-of-the-art performance on widely used benchmarks. The source code is available at \href{https://github.com/marvineta0405/DyGCN.git}{GitHub}.
\end{abstract}

\section{Introduction}
\label{sec:intro}
As a fundamental problem in computer vision and deep learning, 
Human Action Recognition (HAR) has attracted increasing attention due to its wide-ranging applications in domains such as surveillance, sports analytics, and behavioral 
assessment~\cite{sign, iMiGUE, SKI, pedestrian}. However, early HAR methods primarily relied on image or video data, making them sensitive to illumination changes, background noise, and viewpoint variations. To address these limitations, skeleton-based action recognition methods have been widely adopted due to their computational efficiency and robustness to environmental noise.\\
\indent Early approaches to skeleton-based action recognition mainly relied on Convolutional Neural Networks (CNNs) and Recurrent Neural Networks (RNNs). CNN-based methods \cite{ke2017,kim2017} transform skeleton sequences into image-like representations, whereas RNN-based methods \cite{yongdu2015,wang2017,liu2018} model long-term dependencies among joints. However, both paradigms largely overlook the intrinsic connectivity and motion structure of the human body. For that reason, Graph Convolutional Networks (GCNs) have recently become the dominant approach for modeling human joint dynamics from skeleton sequences \cite{stgcn,ctrgcn,stagcn,msg3d}. By representing the human skeleton as a graph, where joints are treated as nodes and bones as edges, GCN-based methods can effectively capture spatial-temporal dependencies and complex inter-joint interactions for accurate action analysis. Despite their effectiveness, conventional GCN-based methods are fundamentally limited by the adjacency matrix, which only models pairwise joint connections, thereby failing to capture multi-joint dependencies and neglecting high-order correlations essential for recognizing subtle and complex actions.\\ \indent To bridge this gap, hypergraph learning~\cite{hgnn} was introduced for skeleton-based action recognition. Unlike conventional graphs in GCN-based methods, where edges are restricted to pairwise joint connections through the adjacency matrix, hyperedges are defined via an incidence matrix that connects multiple joints simultaneously, enabling the modeling of high-order relationships among joint groups. Several studies~\cite{selective-HCN,hgnn} construct the incidence matrix based on predefined human anatomical prior knowledge, providing a stable physical skeletal structure. While these methods have demonstrated effectiveness in capturing higher-order relationships among joints, two key limitations remain: (1) fixed joint groupings within hyperedges constrain the model’s flexibility by failing to capture dynamic correlations among joints that vary significantly across action categories; and (2) existing methods often emphasize global structural patterns while overlooking subtle motion details of key body parts, thereby missing fine-grained action-specific relations and making visually similar actions harder to distinguish.\\ \indent To address these limitations, we propose a novel Dynamic Hypergraph Construction Network with Contrastive Score Refinement, namely DyGCT. Specifically, DyGCT consists of two core parts: Geometry-Constrained Dynamic Hypergraph Construction (GeoHC) and Contrastive Score Refinement (CSR). GeoHC is designed to mitigate the limitations of static hypergraph structures by enabling the model to dynamically construct action-specific hyperedges through second-order compatibility, inspired by prior works~\cite{mac1,mac2}. Different from prior HGCN methods that rely on predefined incidence matrices, GeoHC derives hyperedges from learned compatibility scores in an action-specific manner. Moreover, in contrast to~\cite{mac1,mac2}, which define second-order compatibility for homogeneous entities based only on direct first-order relations, GeoHC extends this formulation to heterogeneous skeleton joints by incorporating topology-aware structural bias and introducing a complementary compatibility term to capture indirect joint dependencies. To further promote discriminative features among ambiguous action categories, we propose a Contrastive Score Refinement (CSR) module that explicitly refines the pairwise compatibility score during the hypergraph construction process. Extensive experiments on three benchmark datasets, namely NTU RGB+D, NTU RGB+D 120, and NW-UCLA, demonstrate that DyGCN achieves state-of-the-art performance on widely used benchmarks.
\section{Related Work}
\subsection{GCN-Based Methods}
In recent years, graph-structured data has become increasingly common, as nodes and edges can naturally represent the topological structure of the human skeleton. Accordingly, GCN-based models \cite{ctrgcn,stagcn,2sagcn,msg3d,infogcn,stgcn} have become dominant for skeleton-based action recognition by modeling spatial-temporal dependencies over skeleton graphs. Chen et al. \cite{ctrgcn} proposed CTR-GCN, a channel-wise topology refinement network that learns dynamic, sample-adaptive topologies and fuses joint features across channels. HD-GCN \cite{hdgcn} improves feature extraction via a hierarchical decomposition of joints that models both short-range and long-range interactions. BlockGCN~\cite{blockgcn} encodes bone connectivity through graph distance representations, achieving superior performance over conventional GCNs while substantially reducing model parameters. However, adaptive GCNs still learn topology through pairwise joint edges. In contrast, DyGCN learns which sets of joints should form action-dependent hyperedges, shifting topology learning from pairwise edge refinement to group-level structural inference.
\subsection{HGCN-Based Methods} In contrast to GCNs, which are constrained to pairwise interactions, HGCN-based methods employ hyperedges to encode group-wise joint correlations, enabling high-order structural reasoning. Hao et al.~\cite{hgnn} introduced Hyper-GNN, one of the first works to exploit hypergraph learning for capturing both spatial-temporal dynamics and high-order joint dependencies. Selective-HCN~\cite{selective-HCN} introduces a Selective-scale Hypergraph Convolution module for capturing multi-scale skeleton information and a Selective-frame Temporal Convolution module for adaptively filtering discriminative frames. Wu et al. \cite{sthg} proposed STHG-DAN, which divides the human body into three parts and constructs multi-view spatial-temporal hypergraphs to extract higher-order features for action recognition. KEHCN \cite{KEH} leverages hypergraph convolution to capture
higher-order joint correlations, further enhanced by LLM-based
kinematic supervision and a mixture-of-experts gating module for
comprehensive motion representation. A limitation of the mentioned approaches is their reliance on fixed hypergraph structures derived from human prior knowledge, overlooking the dynamic action-specific joint correlations. 
\subsection{Contrastive Learning} Recently, contrastive learning has made significant advances across a wide range of fields \cite{moco, simclr}. In action recognition scenarios, \cite{hico,crossclr,actionlet} employ contrastive learning to learn invariant representations from skeleton data in an unsupervised manner. Under supervised learning, several approaches~\cite{skeletonGCL,frhead,thtformer} leverage contrastive learning to refine skeleton representations and enhance model performance. Specifically, skeletonGCL~\cite{skeletonGCL} explicitly exploits rich cross-sequence relations through graph contrastive learning. FR-Head~\cite{frhead} incorporates multi-level contrastive learning into GCN-based methods to improve discrimination between ambiguous actions. THT-Former~\cite{thtformer} integrates contrastive learning within a Topology-adaptive Hypergraph Transformer Network to refine local action representations and obtain more discriminative skeleton features. Different from THT-Former, which mainly refines skeleton representations within a transformer-based hypergraph framework, our DyGCN explicitly learns action-specific hypergraph topology by dynamically constructing the incidence matrix from compatibility scores. Thus, contrastive learning in our method is applied to refine the compatibility representation used for hypergraph construction, rather than directly supervising local action features.
\section{Method} 
\subsection{Preliminaries} 
\noindent\textbf{Graph Convolution.} A human action can be described as a series of skeleton graphs, where each skeleton is formulated as a graph $G = (\mathcal{V}, \mathcal{E})$. In this representation, joints correspond to vertices in the set $\mathcal{V} = \{v_1, v_2, \ldots, v_N\}$, while bones define the edges $\mathcal{E}$ connecting these vertices. The input skeleton sequence is denoted as a matrix $\mathbf{X} \in \mathbb{R}^{T \times V \times C}$, where $T$ denotes the number of frames, $V$ represents the number of vertices, and $C$ indicates the feature dimension of each vertex. The conventional Graph Convolution (GCN) process for an input skeleton feature X at layer k can be mathematically represented as follows:
\begin{equation}
\mathbf{F}^k = \sigma \left( \tilde{\mathbf{D}}^{-1/2} \tilde{\mathbf{A}} \tilde{\mathbf{D}}^{-1/2} \mathbf{X}^k \mathbf{W}^{k}  \right) .
\end{equation}
where $\sigma(\cdot)$ denotes the nonlinear activation function ReLU. $\tilde{\mathbf{A}} \in \mathbb{R}^{V \times V}$ is the adjacency matrix with added self-loops, where $V$ represents the number of human joints. $\tilde{\mathbf{D}}$ is the degree matrix of $\tilde{\mathbf{A}}$.   $\mathbf{W}^{k} \in \mathbb{R}^{C^k \times C^{k+1}}$ is the learnable weight matrix utilized for feature transformation in the feature space.\\
\noindent\textbf{Hypergraph Convolution.} Unlike standard graphs that model only pairwise relationships via edges, hypergraphs use hyperedges to connect multiple vertices simultaneously, enabling the model to capture high-order relational structure. Generally, a hypergraph for human skeleton is defined as $\mathcal{H} = (\mathcal{V}_h, \mathcal{E}_h,\mathcal{W}_h)$, where $\mathcal{V}_h$ and $\mathcal{E}_h$ denote the sets of vertices and hyperedges. In addition, $\mathcal{W}_h$ represents the weight matrix, which encodes the strength of each hyperedge. To represent the multi-vertex interactions induced by hyperedges, the hypergraph $\mathcal{H}$ can be denoted by a $|\mathcal{V}| \times |\mathcal{E}|$ incidence matrix $H \in \mathbb{R}^{V \times E}$. V  indicates the number of joints and E indicates the number
of hyperedges. In detail, with $v \in \mathcal{V}_h$ and $e \in \mathcal{E}_h$, the elements of the incidence matrix are determined by:
\begin{equation}
h(v,e)=
\begin{cases}
1, & v\in e\\
0, & v\notin e
\end{cases}
\end{equation}
In a hypergraph, the notion of degree is defined by both vertices (joints) and hyperedges. For a joint $v \in \mathcal{V}_h$, its degree is computed as the total weight of the hyperedges that contain $v$, expressed as
\begin{equation}
d(v) = \sum_{e \in \mathcal{E}_h} \mathcal{W}_h(e)\, h(v,e) ,
\end{equation}

where $\mathcal{W}_h(e)$ denotes the weight of hyperedge e. The degree of a hyperedge $e \in \mathcal{E}_h$  \begin{figure*}[t]
    \centering
    \includegraphics[width=0.9\textwidth]{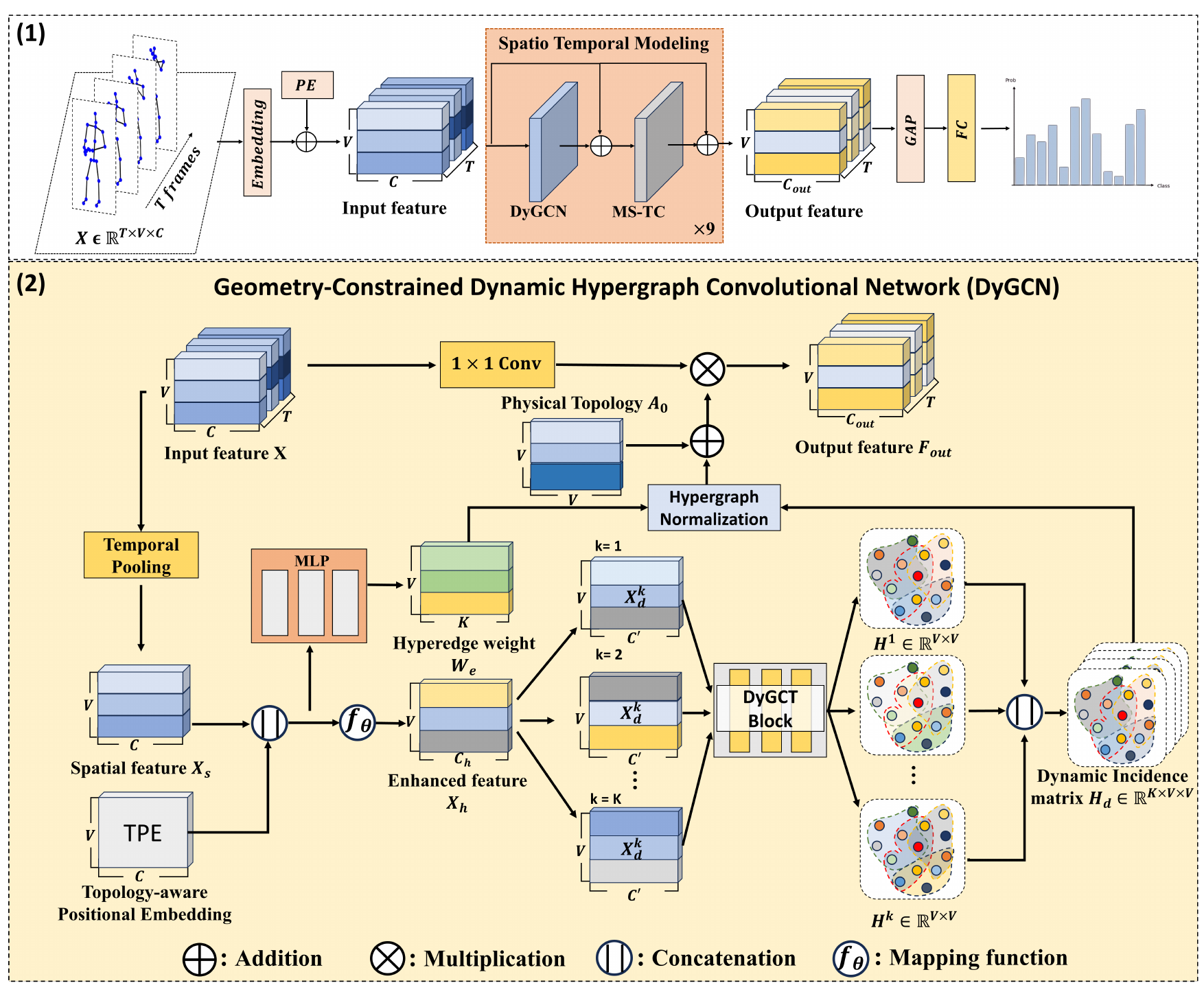}
    \caption{Overview of our proposed method.  (1) The overall pipeline takes skeleton sequences $\mathbf{X}$ as input, followed by embedding, spatio-temporal modeling via stacked DyGCN and MS-TC \cite{mstc} modules, and global average pooling with Fully Connected Layers (FC) for action classification. (2) The detailed architecture of the 
Dynamic Hypergraph Convolutional Network (DyGCN)}
    \label{fig_overview}
\end{figure*} \\is defined as the number of vertices contained in that hyperedge, given by
\begin{equation}
d(e) = \sum_{v \in \mathcal{V}_h} h(v,e).
\end{equation}
To represent the degrees of joints and hyperedges, diagonal matrices
$D_v \in \mathbb{R}^{V \times V}$ and $D_e \in \mathbb{R}^{E \times E}$
are constructed, whose diagonal entries correspond to the degrees of vertices
and hyperedges, respectively. Here, the Hypergraph Convolution \cite{hcn} applied to the input skeleton feature X at layer L can be written as :
\begin{equation}
\label{eq:hyper}
X^{(L+1)} = \sigma\left(D_v^{-1} H W D_e^{-1} H^\top X^{(L)} \Theta^{(L)}\right),
\end{equation}
where $W$ is a diagonal matrix representing the hyperedge weights, and $\Theta^{(L)} \in \mathbb{R}^{F^{(L)} \times F^{(L+1)}}$ is the learnable matrix between the $L$-th and $(L+1)$-th layers.
\subsection{Dynamic Hypergraph Convolution Module} 
\begin{figure*}[t]
    \centering
    \includegraphics[width=0.9\textwidth]{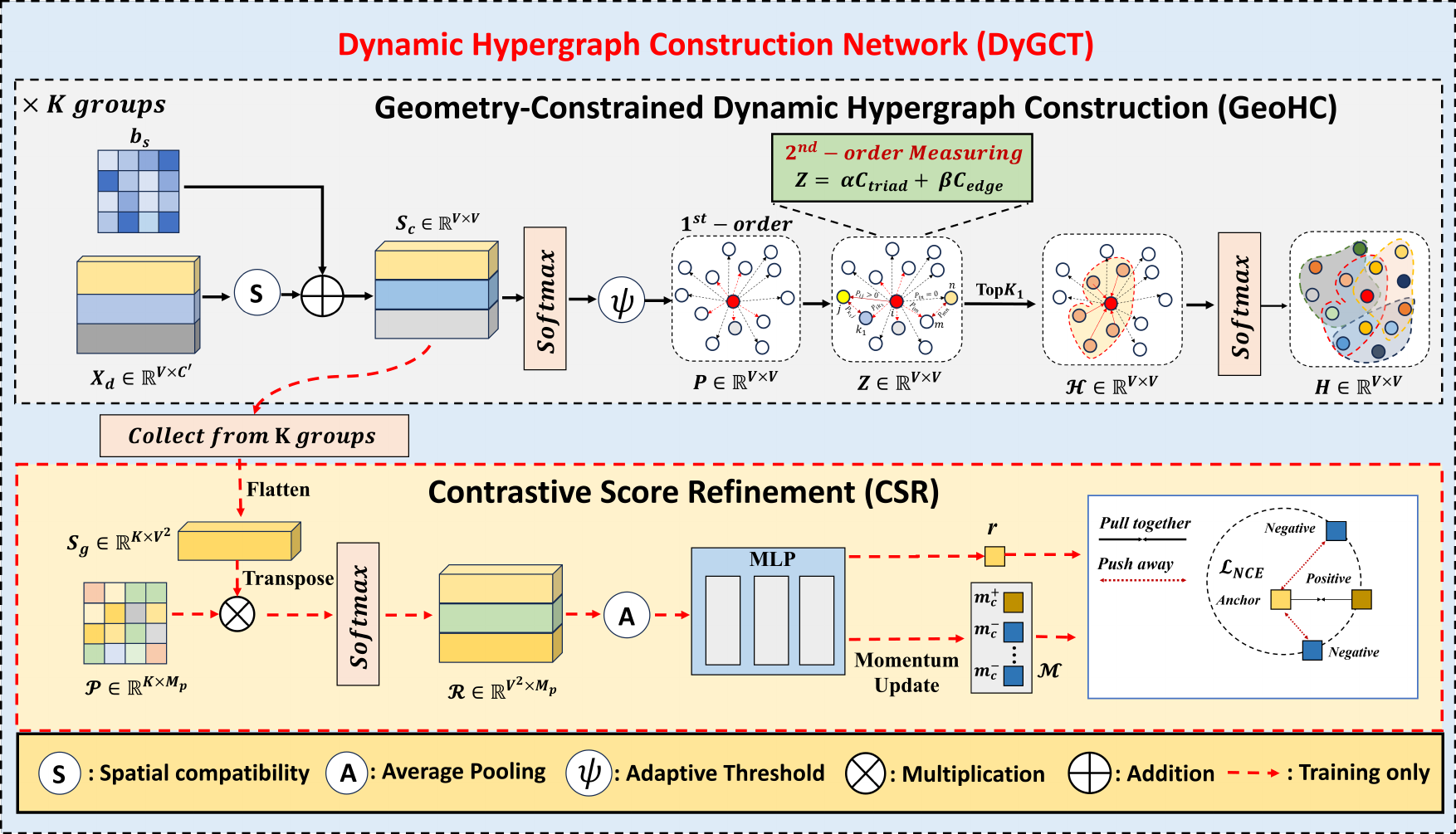}
    \caption{The framework of the DyGCT block, comprising the Geometry-Constrained Dynamic Hypergraph Construction (GeoHC) module and the Contrastive Score Refinement (CSR) module.}
    \label{dygct}
\end{figure*}

In this subsection, we first introduce the GeoHC module to construct the dynamic hypergraph incidence matrix. Based on the constructed hypergraph, DyGCN is applied to capture high-order joint interactions. The overall architecture is shown in Fig.~\ref{fig_overview}.\\
\textbf{Hypergraph Construction.} The construction of the hypergraph incidence matrix is critical for modeling high-order structural dependencies. Existing approaches typically rely on a static hypergraph derived from a natural fixed topological structure, which constrains the model’s ability to capture action-dependent relationships among vertices across different actions. To address this limitation, we propose a strategy for dynamically constructing the incidence matrix for hypergraph convolution, namely Geometry-Constrained Dynamic Hypergraph Construction (GeoHC), as shown in Fig.~\ref{dygct}. \\
 \indent In the proposed framework, we construct hyperedges by treating each joint as the  Center of Mass (CoM), resulting in a total of $V$ hyperedges. Given the input feature  $\mathbf{X} \in \mathbb{R}^{T \times V \times C}$, we first apply temporal average pooling to obtain a spatial representation $\mathbf{X}_S \in \mathbb{R}^{V \times C}$, where $V$ represents the number of joints and $C$ is the input channel dimension. To preserve anatomical structure during dynamic hypergraph construction, we introduce a Topology-aware Positional Embedding (TPE) based on skeleton hop distances, inspired by RPE~\cite{hyperformer}. Unlike the original formulation that biases pairwise attention scores, we aggregate hop-distance encodings across all joints to produce a per-node topology-aware embedding.
Specifically, let $D(i,j)$ denote the hop distance between joint $i$ and joint $j$ in the predefined skeleton topology, we define a learnable embedding table $\mathbf{E} \in \mathbb{R}^{(D_{\max}+1) \times C}$ to map each discrete hop distance to a feature vector. 
The structural embedding $\mathbf{p}_i $ of joint $i$ is obtained by aggregating the relative embeddings with respect to all other joints:
\begin{equation}
\mathbf{p}_i = \frac{1}{V} \sum_{j=1}^{V} \mathbf{E}_{D(i,j)},
\end{equation}
where $D_{\max}$ denotes the maximum hop distance. The enriched node representation is formed by concatenating the spatial feature and the structural embedding:
\begin{equation}
\tilde{\mathbf{X}}_S(i) = \left[ \mathbf{X}_S(i) \,\|\, \mathbf{p}_i \right].
\end{equation}
This design allows each joint to encode not only its own appearance information but also its relative structural position within the skeleton graph, thereby enhancing the modeling of topology-aware interactions before dynamic hyperedge construction. Then, we project the resulting representation into a hidden feature dimension through a mapping function $f_{\theta}: \mathbb{R}^{2C} \rightarrow \mathbb{R}^{C_h}$, obtaining $\mathbf{X}_h \in \mathbb{R}^{V \times C_h}$. Following~\cite{blockgcn}, we split the projected representation $\mathbf{X}_h$ into $\{\mathbf{X}_d^{(k)}\}_{k=1}^{K}$, where each group $\mathbf{X}_d^{(k)} \in \mathbb{R}^{V \times C'}$ captures joint differences in a distinct semantic subspace and $C'=\lfloor C_h/K \rfloor$. This grouped design encourages the model to learn diverse relational patterns across different semantic subspaces. Since the following compatibility construction is shared across groups, we denote a representative group as $\mathbf{X}_d\in\mathbb{R}^{V\times C'}$ for simplicity.\\
\indent While Euclidean distance is commonly utilized to measure joint correspondence, first-order feature distances alone are insufficient to capture the complex high-order dependencies among skeleton joints. Motivated by the second-order compatibility measure proposed for point cloud registration in~\cite{mac1}, we adapt this concept to the skeleton domain by extending it to capture high-order joint interactions for action recognition. Specifically, we first construct a pairwise compatibility score between joints based on their distance in the projected feature space. However, the compatibility score in~\cite{mac1} is defined for homogeneous entities, whereas skeleton joints are inherently heterogeneous --- each joint plays a unique anatomical role in the human body. To address this, we augment the joint compatibility score with a learned hop-distance bias to encode structural relationships between joints. Let $\mathbf{q}_i$ and $\mathbf{q}_j$ be the $i$-th and $j$-th row vectors of $\mathbf{X}_d \in \mathbb{R}^{V \times C'}$, the entries of the pairwise compatibility matrix $\mathbf{S}_c$ are defined as:

\begin{equation}
S_{i,j}
=
\underbrace{
-\frac{\|\mathbf{q}_i-\mathbf{q}_j\|_2^2}{2\sigma^2}
}_{\text{Spatial compatibility}}
+
\underbrace{
b_{D(i,j)}
}_{\text{Hop-distance bias}}.
\end{equation}

where $S_{i,j}$ denotes the $(i,j)$-th entry of the pairwise compatibility matrix $\mathbf{S}_c$, representing the compatibility score between joints $i$ and $j$. $\|\mathbf{q}_i - \mathbf{q}_j\|_2^2$ denotes the squared Euclidean distance between $\mathbf{q}_i$ and $\mathbf{q}_j$, $\sigma$ is a learnable distance parameter, and $b_{D(i,j)}$ is a learned structural bias indexed by the shortest-path hop distance $D(i,j)$ between joints $i$ and $j$.  The first-order affinity matrix $\mathbf{F}$ is then constructed by applying row-wise softmax to the pairwise compatibility scores with self-connections explicitly removed, expressed as:
\begin{equation}
F_{i,j} = \frac{\exp(S_{i,j})}{\sum_{l \neq i} \exp(S_{i,l})}, \quad F_{i,i} = 0
\end{equation}
Based on this affinity measure, two joints are considered compatible when $F_{i,j}$ exceeds a threshold $\psi$. However, rather than using a 
static global threshold as in~\cite{mac1}, we introduce an adaptive threshold $\psi_i$ for each joint as the mean of its $\operatorname{TopK_0}$ affinity values:

\begin{equation}
\psi_i = \frac{1}{K_0} \sum_{t \in \mathcal{T}_i} F_{i,t}, 
\quad \text{where } \mathcal{T}_i = \operatorname{TopK_0}\left(\{F_{i,j}\}_{j=1}^{V}\right),
\end{equation}
where $K_0$ is a hyperparameter denoting the number of entries selected for computing the adaptive threshold $\psi_i$. Subsequently, the first-order compatibility matrix $\mathbf{P}$ is defined as:
\begin{equation}
\quad P_{i,j} = \begin{cases} F_{i,j}, & \text{if } F_{i,j} \geq \psi_i \\ 0, & \text{if } F_{i,j} < \psi_i \end{cases} ,
\end{equation}
where $P_{i,j}$ denotes the $(i,j)$-th entry of $\mathbf{P}$. To further capture high-order relational structures, similar 
to~\cite{mac1}, the second-order spatial compatibility $C_{\mathrm{triad}}$ is defined as the sum of weighted products over common compatible joints shared between joints $i$ and $j$, 
formulated as:
\begin{equation}
\quad C^{\text{triad}}_{i,j} = P_{i,j} \cdot \sum_{l=1}^{V} P_{i,l} \cdot P_{l,j},
\end{equation}
 where $C^{\text{triad}}_{i,j}$ denotes the $(i,j)$-th entry of $\mathbf{C}_{\text{triad}}$, representing the second-order compatibility score between joints $i$ and $j$. However, this formulation requires joints $i$ and $j$ to have non-zero first-order compatibility, i.e., $P_{i,j} > 0$, and therefore ignores potentially meaningful indirect relationships mediated by shared compatible joints. As a result, the hypergraph may fail to select joints without direct first-order compatibility as valid hyperedge members during hyperedge construction. To alleviate this limitation, we extend the second-order spatial compatibility $\mathbf{C}_{\text{triad}}$ with a complementary term $\mathbf{C}_{\text{edge}}$ to incorporate indirect joint relationships bridged by shared intermediate joints. The overall compatibility score is then formulated as: 
\begin{equation}
C^{\text{edge}}_{i,j} = 
\left(1 - \mathbbm{1}[P_{i,j} > 0]\right)
\cdot
\sum_{l=1}^{V} P_{i,l} P_{l,j},
\end{equation}
\begin{equation}
\mathbf{Z} = \alpha \cdot\mathbf{C}_{\text{triad}} + \beta \cdot\mathbf{C}_{\text{edge}},
\end{equation}
where $C^{\text{edge}}_{i,j}$ denotes the $(i,j)$-th entry of $\mathbf{C}_{\text{edge}}$, capturing indirect joint relationships through shared intermediate joints. $\alpha$ and $\beta $ are learnable weights balancing the contributions of both terms in the final second-order compatibility matrix $\mathbf{Z}$. \\
\indent To encourage the model to capture action-specific features, we 
constrain hyperedge membership to the most semantically compatible 
joints rather than assigning all joints to every hyperedge. Given 
each row vector $\mathbf{z}_i$ in the compatibility matrix 
$\mathbf{Z}$, which represents the compatibility score between 
joint $i$ and all other joints, we select the $\operatorname{TopK_1}$ most compatible joints as the set of hyperedge members $\mathcal{N}_i$:
\begin{equation}
\mathcal{N}_i = \operatorname{TopK_1}(\mathbf{z}_i),
\end{equation}
where $K_1$ is a hyperparameter controlling the number of joints 
in each hyperedge. The selected compatibility scores in $\mathcal{N}_i$ are then normalized via softmax to estimate the hyperedge membership probability, constructing the incidence matrix $\mathbf{H} \in \mathbb{R}^{V \times E}$ for the representative group,  where $E=V$ since each joint is treated as a hyperedge center: 
\begin{equation}
\quad H_{j,i} =
\begin{cases}
\dfrac{\exp(z_{i,j})}
{\sum_{l \in \mathcal{N}_i} \exp(z_{i,l})},
& j \in \mathcal{N}_i, \\
0, & j \notin \mathcal{N}_i,
\end{cases}
\end{equation}
Since the same construction is applied independently to each group, we obtain group-specific incidence matrices $\{\mathbf{H}^{(k)}\}_{k=1}^{K}$. The resulting incidence matrices $\mathbf{H}^{(k)}$ from all $K$ groups are then stacked along the group dimension to form the final dynamic incidence tensor:
\begin{equation}
\mathbf{H}_d =
\operatorname{Stack}_{k=1}^{K} \mathbf{H}^{(k)}
\in \mathbb{R}^{K \times V \times V}.
\end{equation}
\textbf{DyGCN.} In Eq.~\ref{eq:hyper}, the hypergraph convolution is determined by both the incidence matrix and the hyperedge weight matrix. Therefore, we employ a learned MLP layer to estimate the hyperedge weights $\mathbf{W}_e  \in \mathbb{R}^{K \times V}$ based on the enriched node representation $\tilde{\mathbf{X}}_S$. 
\begin{equation}
\mathbf{W}_e = \sigma_2 \left( \mathrm{MLP}\!\left(\tilde{\mathbf{X}}_S\right)\right), 
\end{equation}
where $\sigma_2(\cdot)$ denotes the nonlinear activation function Tanh. For hypergraph normalization, $\mathbf{W}_e$ is lifted to a group-wise diagonal tensor $\mathcal{D}(\mathbf{W}_e) \in \mathbb{R}^{K \times V \times V}$. The degree matrices are computed separately for each group and then stacked along the group dimension, yielding $D_v,D_e \in \mathbb{R}^{K \times V \times V}$. Subsequently, the hypergraph normalization $\tilde{\mathbf{H}}_d$ induced by the dynamic incidence tensor $\mathbf{H}_d$ and the hyperedge weights is defined as follows:
\begin{equation}
    D_v^{(k)} = \operatorname{diag}\left(\mathbf{H}^{(k)}w_e^{(k)}\right),
    \quad
    D_e^{(k)} = \operatorname{diag}\left((\mathbf{H}^{(k)})^\top \mathbf{1}_V\right).
\end{equation}
\begin{equation}
     \tilde{\mathbf{H}}_{d}^{(k)}
    =
    (D_v^{(k)}+\epsilon I_V)^{-1}
    \mathbf{H}^{(k)}
    \mathcal{D}(w_e^{(k)})
    (D_e^{(k)}+\epsilon I_E)^{-1}
    (\mathbf{H}^{(k)})^\top,    
\end{equation}
\begin{equation}
    \tilde{\mathbf{H}}_{d}
    =
    \operatorname{Stack}_{k=1}^{K} \left(\tilde{\mathbf{H}}_{d}^{(k)}\right)
    \in \mathbb{R}^{K \times V \times V}.
\end{equation}

where $w_e^{(k)} \in \mathbb{R}^{E}$ denotes the hyperedge weight vector for the $k$-th group, and the normalization is performed independently for each group, yielding $\tilde{\mathbf{H}}_d \in \mathbb{R}^{K \times V \times V}$. Furthermore, we incorporate the physical topology $A_0$ in order to preserve the inherent physical connectivity of the skeleton. Specifically, following~\cite{stgcn}, which partitions the physical topology of the human body into three subsets, namely centripetal (cp), identity (id), and centrifugal (cf), we construct $A_0$ by combining these subsets into a unified adjacency matrix and repeating it across the $K$ groups, yielding $A_0 \in \mathbb{R}^{K \times V \times V}$. Let $\mathbf{X}^{(L)} \in \mathbb{R}^{T \times V \times C_L}$ denote the input feature of the $L$-th DyGCN layer. The input feature $\mathbf{X}^{(L)}$ is split into $K$ channel groups, i.e., $\mathbf{X}^{(L)} = \operatorname{Concat}_{c}(\mathbf{X}_{1}^{(L)}, \mathbf{X}_{2}^{(L)}, ..., \mathbf{X}_{K}^{(L)})$, where $\mathbf{X}_{k}^{(L)} \in \mathbb{R}^{T \times V \times C_L / K}$. The group-specific transformation matrix is defined as $\Theta_k^{(L)} \in \mathbb{R}^{C_L / K \times C_{L+1} / K}$. The spatial propagation matrix $A_0^{(k)}+\lambda \tilde{\mathbf H}_d^{(k)} \in \mathbb{R}^{V \times V}$ is applied along the joint dimension and shared across all temporal steps, and the outputs from all groups are concatenated along the channel dimension. Then, the hypergraph convolution in Eq.~\ref{eq:hyper} can be rewritten as:
\begin{equation}
\mathbf{X}^{(L+1)}
=
\sigma
\left(
\operatorname*{\|}_{k=1}^{K}
\left[
\left(
A_0^{(k)}
+
\lambda \tilde{\mathbf H}_d^{(k)}
\right)
\mathbf{X}_k^{(L)}
\Theta_k^{(L)}
\right]
\right),
\end{equation}
where $\mathbf{X}_k^{(L)}$ denotes the $k$-th channel partition of $\mathbf{X}^{(L)}$, $\operatorname*{\|}$ denotes channel-wise concatenation, and $\lambda$ is a learnable parameter controlling the contribution of the dynamic hypergraph topology relative to the physical topology.

\subsection{Contrastive Score Refinement}
To encourage the dynamic hypergraph construction to capture fine-grained action-specific relational structures, we propose Contrastive Score Refinement (CSR) to refine the learned compatibility scores, as shown in Fig.~\ref{dygct}. Specifically, the compatibility scores from all $K$ groups are collected and flattened into $\mathbf{S}_g \in \mathbb{R}^{K \times V^2}$. We then define a learnable prototype matrix $\mathcal{P}=\{\mathbf{p}_m\in\mathbb{R}^{K}\}_{m=1}^{M_p}$, where each randomly initialized prototype captures a compatibility pattern across the $K$ groups and $M_p$ denotes the number of prototypes. The similarity between the joint-pair compatibility representation $\mathbf{S}_g$ and $\mathcal{P}$ is computed via inner product followed by softmax normalization, formulated as:
\begin{equation}
\mathcal{R} = \operatorname{softmax}\left(\mathbf{S}_g^\top 
\mathcal{P}\right) \in \mathbb{R}^{V^2 \times M_p},
\end{equation}
where $\mathcal{R}_{i,m}$ denotes the $(i,m)$-th entry of $\mathcal{R}$,
reflecting the similarity between the $i$-th joint-pair compatibility
representation and the $m$-th prototype. Subsequently, $\mathcal{R}$ is aggregated along the prototype dimension via average pooling and projected via an MLP to obtain the compatibility representation vector $\mathbf{r}$:

\begin{equation}\mathbf{r} = \text{MLP}\left(\text{AvgPool}(\mathcal{R})
\right) \in \mathbb{R}^{C_g}
\end{equation}

Consequently, to further leverage discriminative cross-sequence 
context to distinguish similar action patterns, we adopt 
contrastive loss from~\cite{moco, skeletonGCL} to supervise 
the representation vector $\mathbf{r}$. Given the memory 
bank $\mathcal{M}$ and a batch of representation 
samples, the momentum update is defined as:

\begin{equation}
\mathbf{m}_c \leftarrow \mu \cdot \mathbf{m}_c + (1 - \mu) \cdot \mathbf{r},
\end{equation}

where $\mathbf{m}_c \in \mathcal{M}$ is the compatibility 
representation entry for class $c$ and $0 < \mu < 1$ is a 
momentum coefficient. The corresponding contrastive loss is 
formulated as:

\begin{equation}
\label{eq:nce}
\mathcal{L}_{\mathrm{C}} =
- \log
\frac{
\exp\!\left(\mathrm{sim}(\mathbf{r}, \mathbf{m}^+_c) / \tau\right)
}{
\exp\!\left(\mathrm{sim}(\mathbf{r}, \mathbf{m}^+_c) / \tau\right)
+
\sum_{\mathbf{m}^-_c \in \mathcal{M}^{-}}
\exp\!\left(\mathrm{sim}(\mathbf{r}, \mathbf{m}^-_c) / \tau\right)
}
\end{equation}
where $\mathrm{sim}(\cdot, \cdot)$ denotes the cosine similarity 
function, $\mathbf{m}^+_c \in \mathcal{M}$ denotes the memory 
bank entry of the same class as $\mathbf{r}$, 
$\mathbf{m}^-_c \in \mathcal{M}^{-}$ denotes memory bank entries 
of different classes, and $\tau$ is a temperature parameter.
In general, the overall training objective of the proposed method is formulated as the sum of multiple loss terms, which can be written as: 
\begin{equation}
\mathcal{L} = \mathcal{L}_{CE} + \gamma\cdot \mathcal{L}_{\mathrm{C}} 
\end{equation}
where $\mathcal{L}_{CE}$ denotes the cross-entropy loss for supervising action classification. $\gamma$ is the hyperparameter to control the Contrastive Score Refinement loss $\mathcal{L}_{\mathrm{C}}$. Notably, the Contrastive Score Refinement loss $\mathcal{L}_{\mathrm{C}}$ is employed only during training and omitted during inference, ensuring that the proposed framework incurs no additional computational overhead.
\section{Experiments} 
\begin{table*}[t]
\centering
\caption{Comparison with state-of-the-art methods on NTU RGB+D \cite{ntu}, NTU RGB+D 120 \cite{ntu120}, and NW-UCLA \cite{ucla}. * denotes methods using additional supervision signals or non-standard input data.}
\label{tab:sota}
\resizebox{\textwidth}{!}{
\begin{tabular}{l|l|c|cc|cc|c}
\hline
\multirow{2}{*}{Types} & \multirow{2}{*}{Methods} & \multirow{2}{*}{Publication} & \multicolumn{2}{c|}{NTU RGB+D} & \multicolumn{2}{c|}{NTU RGB+D 120} & NW-UCLA \\
\cline{4-8}
& & & X-Sub(\%) & X-View(\%) & X-Sub(\%) & X-Set(\%) & Top-1(\%) \\
\hline
\multirow{10}{*}{GCN-based}
& ST-GCN \cite{stgcn}        & AAAI2018    & 81.5 & 88.3 & 70.7 & 73.2 & - \\
& 2s-AGCN \cite{2sagcn}        & CVPR2019    & 88.5  & 95.1 & 82.5 & 84.2 & - \\
& CTR-GCN \cite{ctrgcn}        & ICCV2021   & 92.4 & 96.8 & 88.9 & 90.6 & 96.5 \\
& EfficientGCN-B4 \cite{b4} & TPAMI2022  & 91.7 & 95.7 & 88.3 & 89.1 & 97.0 \\
& InfoGCN \cite{infogcn}       & CVPR2022   & 93.0 & 97.1 & 89.8 & 91.2 & 97.0 \\
& GAP* \cite{gap}            & ICCV2023   & 92.9 & 97.0 & 89.9 & 91.1 & 97.2 \\
& HD-GCN*(6s) \cite{hdgcn}     & ICCV2023   & 93.4 & 97.2 & 90.1 & 91.6 & 97.2 \\
& BlockGCN \cite{blockgcn}       & CVPR2024   & 93.1 & 97.0 & 90.3 & 91.5 & 96.9 \\
& HiOD   \cite{hiod}      & ICCV2025   & 93.6 & 97.4  & 90.2 & 91.6 & - \\
& ProtoGCN  \cite{protogcn}      & CVPR2025   & 93.5 & 97.5 & 90.4 & 91.9 & 96.9 \\
\hline
\multirow{4}{*}{Transformer-based}
& DSTA-Net \cite{dsta} & ACCV2020    & 89.5 & 95.7 & 86.6    & 89.0    & - \\
& STST\cite{stst}    & ACMM2021   & 90.8 & 96.6 & -    & -    & - \\
& Skeleton MixFormer\cite{mixformer}    & ACMM2023   & 93.0 & 97.0 & 90.0    & 91.3    & 97.2 \\
& THTFormer \cite{thtformer}    & PR2025   & 93.0 & 96.8 & 90.1    & 91.3    & 97.1 \\
\hline
\multirow{4}{*}{HGNN-based}
& Hyper-GNN \cite{hgnn}       & TIP2021     & 89.5 & 95.7 & -    & -    & - \\
& Selective-HCN \cite{selective-HCN}  & ICMR2022   & 90.8 & 96.6 & -    & -    & - \\
& STHG-DAN \cite{sthg}       & PR2024    & 91.2 & 96.5 & 88.7 & 89.8 & - \\
& KEHCN* \cite{KEH}       & ACMM2025    & 93.5 & 97.3 & 90.4& 91.8 & \textbf{97.4} \\
\hline
\rowcolor{green!15}
\textbf{Ours} & \textbf{DyGCN} & - & \textbf{93.7} & \textbf{97.6} & \textbf{90.8} & \textbf{92.0} & \textbf{97.4} \\
\hline
\end{tabular}
}
\end{table*}
\subsection{Experimental Settings}
Extensive experiments are conducted on three widely used action recognition datasets: \textbf{NTU RGB+D }\cite{ntu}, \textbf{NTU RGB+D 120} \cite{ntu120} and \textbf{NW-UCLA} \cite{ucla}. The details of each dataset can be found in the supplementary material. All experiments are performed using a single NVIDIA RTX A6000 GPU within the PyTorch framework and CUDA 11.4 environment. The models are optimized using stochastic gradient descent (SGD) with a momentum of 0.9 and a weight decay of 0.0004. To improve training stability, a warm-up strategy following \cite{resnet} is applied during the first five epochs. For the NTU RGB+D and NTU RGB+D 120 datasets, training is conducted with a batch size of 64 over 140 epochs. The initial learning rate is set to 0.05 and is reduced by a factor of 10 at the 110th and 120th epochs. For the NW-UCLA dataset, the training configuration uses a batch size of 16. The hyperparameter $K_0$ is \begin{table}[t]
\caption{Ablation study on the contribution of each component in DyGCT under the NTU RGB+D X-Sub setting with the joint input modality only.}
\label{tab:abl1}
\centering
\begin{tabular}{c   c c c   c c c}
\hline
\multirow{2}{*}{Method} & \multicolumn{3}{c}{DyGCT} & \multirow{2}{*}{Params (M)} & \multirow{2}{*}{GFLOPs (G)} & \multirow{2}{*}{Acc (\%)}\\
\cline{2-4}
 & TPE & CM & CSR &  &  \\
\hline
Baseline & -- & -- & -- & 1.5 & 1.97 &89.8 \\
\hline
\multirow{3}{*}{DyGCN} 
& -- & $\checkmark$ & -- & 2.24 & 2.7& 91.2 ($\uparrow$ 1.4) \\
& $\checkmark$ & $\checkmark$ & -- & 2.28 & 2.7 &91.7 ($\uparrow$ 1.9) \\
& $\checkmark$ & $\checkmark$ & $\checkmark$ & 2.3 & 2.7 &\cellcolor{green!15}\textbf{92.0 ($\uparrow$ 2.2)} \\
\hline
\end{tabular}
\end{table}set to $2$ throughout all experiments. The number of groups $K$ is set to 8. Regarding the contrastive learning setup, the momentum coefficient $\mu$ is set to 0.9, and the compatibility representation dimension $C_g$ is set to 256. The temperature parameter $\tau$ is fixed at 0.8 throughout the experiments. We adopt the data preprocessing procedures outlined in CTR-GCN~\cite{ctrgcn} and utilize its architecture as the baseline model.
\subsection{Comparison with State-of-the-Arts}
In this section, we compare our method with state-of-the-art approaches on three benchmark datasets: NTU RGB+D~\cite{ntu}, NTU RGB+D 120~\cite{ntu120}, and NW-UCLA~\cite{ucla}.  For a fair comparison, we follow the same experimental framework as~\cite{ctrgcn}. To comprehensively evaluate DyGCN, we adopt a 4-stream fusion strategy with four input modalities: joint, bone, joint motion, and bone motion. The fused score is obtained by summing the softmax outputs across all four streams. The quantitative results are presented in Table~\ref{tab:sota}. \\
\indent Specifically, DyGCN surpasses all state-of-the-art methods on two large-scale benchmarks: NTU RGB+D  and NTU RGB+D 120. Taking the NTU RGB+D 120 dataset as an example, our method surpasses the most recent HGNN-based method, KEHCN, by margins of 0.4\% on X-Sub and 0.2\% on X-Set. Notably, KEHCN~\cite{KEH} relies on additional pre-trained large language models (GPT-4 and CLIP) to extract fine-grained features, which substantially contribute to its final performance. In addition to evaluations on large-scale datasets such as NTU-60 and NTU-120, we further examine our model on the small-scale NW-UCLA dataset to validate its performance across different data scales. Our method shares the best performance with KEHCN~\cite{KEH} and exceeds the state-of-the-art GCN-based method ProtoGCN~\cite{protogcn} by 0.5\%. Overall, our method consistently outperforms most existing state-of-the-art approaches across different settings, demonstrating its effectiveness. Additional analyses, including computational complexity, runtime comparison, and single-modality performance, are provided in the supplementary material.
\begin{table}[t] 
\centering
\caption{The impact of the hyperparameter $K_1$ in DyGCT (left) and Compatibility Measuring Component $\mathbf{C}_{\text{triad}}$ and $\mathbf{C}_{\text{edge}}$ (right) under NTU RGB+D X-Sub setting.}
\label{tab:abl_combined}
\begin{minipage}[t]{0.5\linewidth}
    \centering
    \begin{tabular}{lccc}
    \hline
    Methods & $K_1$ & w/ CSR & w/o CSR \\
    \hline
    \multirow{4}{*}{DyGCT}
    & 3 & 91.8 & 91.5 \\
    & \cellcolor{green!15}\textbf{5} & \cellcolor{green!15}\textbf{92.0} & \cellcolor{green!15}\textbf{91.7} \\
    & 7 & 91.5 & 91.2 \\
    & 9 & 91.3 & 91.1 \\
    \hline
    \end{tabular}
\end{minipage}
\hfill
\begin{minipage}[t]{0.45\linewidth}
    \centering
    \begin{tabular}{lc}
    \hline
    Methods & Acc (\%) \\
    \hline
    DyGCT & \cellcolor{green!15}\textbf{92.0} \\
    w/o $\mathbf{C}_{\text{triad}}$ & 91.6 \\
    w/o $\mathbf{C}_{\text{edge}}$ & 91.5 \\
    w/o $\mathbf{C}_{\text{triad}}$, $\mathbf{C}_{\text{edge}}$ & 91.4 \\
    \hline
    \end{tabular}
\end{minipage}
\end{table}

\begin{table}[t]
\caption{Performance comparison of DyGCN variants on ambiguous actions.}
\label{tab:underperforming}
\centering
\begin{tabular}{l | ccc | c}
\toprule
 & CTR-GCN\cite{ctrgcn}  & FR-Head\cite{frhead} & HD-GCN\cite{hdgcn} & \textbf{Ours} \\
\midrule
NTU60-Hard      & 62.1 & 67.5 & 67.4 & $\textbf{71.1}^{\text{+3.6}}$ \\
NTU60-Medium    & 83.8 & 83.9 & 84.9 & $\textbf{87.9}^{\text{+3.0}}$ \\
NTU60-Easy      & 95.2 & 95.8 & 95.5 & $\textbf{96.1}^{\text{+0.6}}$ \\
\bottomrule
\end{tabular}
\end{table}
\subsection{Ablation Study}
To examine the effectiveness of each component in the proposed method, we conduct ablation studies on the NTU RGB+D dataset using the joint modality under the X-Sub setting.
\textbf{Effectiveness of Each Module.} In Table~\ref{tab:abl1}, we evaluate the contribution of each component in DyGCT. Using the latest official implementation, the reproduced baseline achieves 89.8\% accuracy. Introducing the proposed Compatibility Measuring (CM) improves the accuracy by 1.4\%, from 89.8\% to 91.2\%. Adding the Topology-aware Positional Embedding (TPE) further yields a 0.5\% gain, showing that GeoHC effectively captures high-order joint relations beyond the pairwise adjacency modeling of the baseline. Built upon GeoHC, the CSR module provides an additional 0.3\% improvement. Although CSR introduces extra training cost, it is used only during training and therefore does not affect inference. Overall, DyGCN improves accuracy from 89.8\% to 92.0\%, corresponding to a 2.2 percentage-point gain, while increasing the model complexity from 1.50M parameters and 1.97 GFLOPs to 2.30M parameters and 2.70 GFLOPs.\\
\textbf{Dynamic Hypergraph Construction.} Since the choice of $K_1$, which controls the maximum number of joints contained in each hyperedge, directly influences the model's recognition accuracy, we present an ablation study in Table~\ref{tab:abl_combined} (left) to examine its impact on  dynamic hypergraph construction under two settings: without Contrastive Score Refinement (CSR) and with CSR. Specifically, a large $K_1$ forces each hyperedge to aggregate noisy joint associations, degrading the quality of the learned representations. In contrast, a small $K_1$ limits the hypergraph's ability to capture complex joint dependencies. As a result, $K_1 = 5$ achieves the best performance under both settings and is therefore adopted for subsequent experiments. Furthermore, the results demonstrate that incorporating CSR consistently yields higher accuracy across all values of $K_1$, highlighting the effectiveness of this module.\\
\textbf{Analysis of Second-order Spatial Compatibility.} We conduct an ablation study to assess the influence of the complementary term $\mathbf{C}_{\text{edge}}$ in the Compatibility Measuring (CM) module and report the results in Table \ref{tab:abl_combined} (right). The variant w/o $\mathbf{C}_{\text{triad}}$, $\mathbf{C}_{\text{edge}}$ represents the case where the CM module relies solely on the first-order affinity matrix $ \mathbf{F}$ for hyperedge construction, without any second-order compatibility reasoning. The results demonstrate that \begin{figure}[t]
    \centering
    \begin{minipage}[t]{0.48\linewidth}
        \centering
        \includegraphics[width=0.65\linewidth]{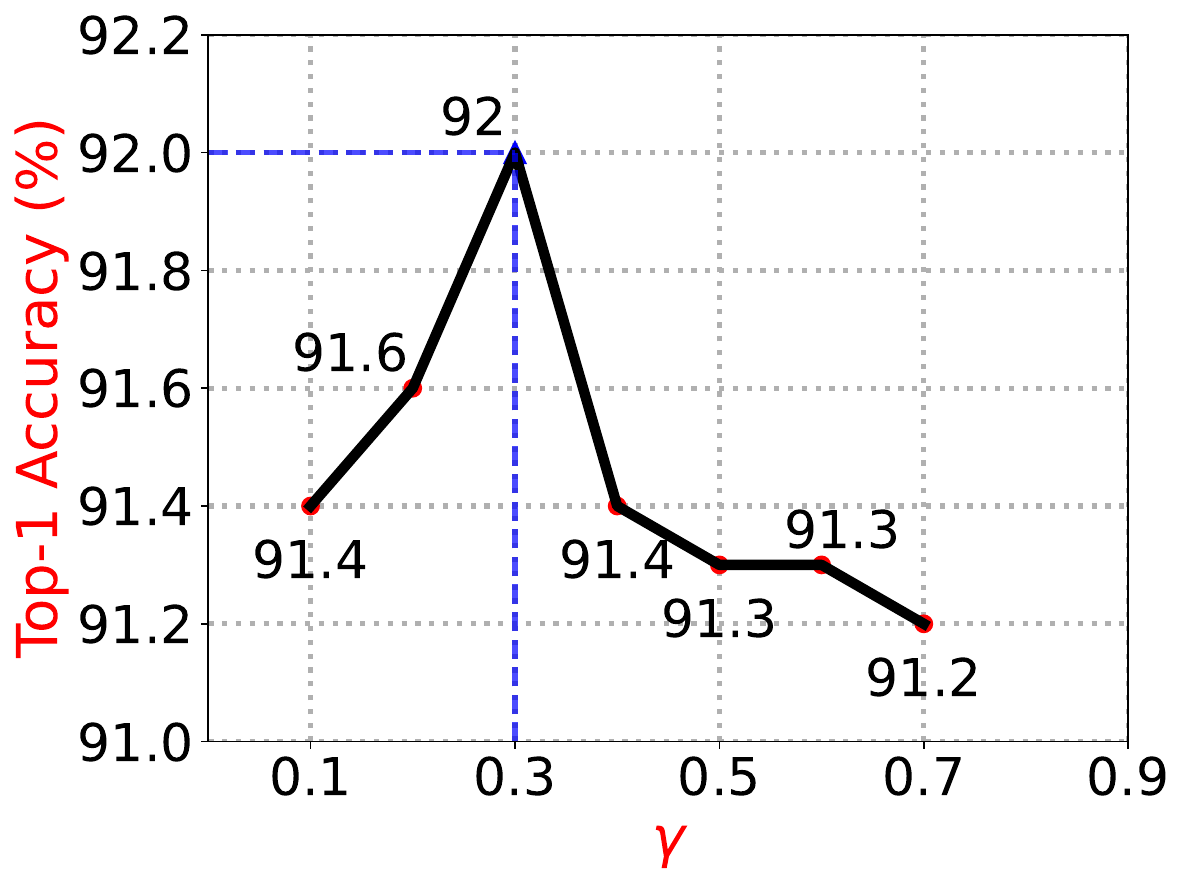}\\
        \vspace{0.5em}
        \includegraphics[width=0.6\linewidth]{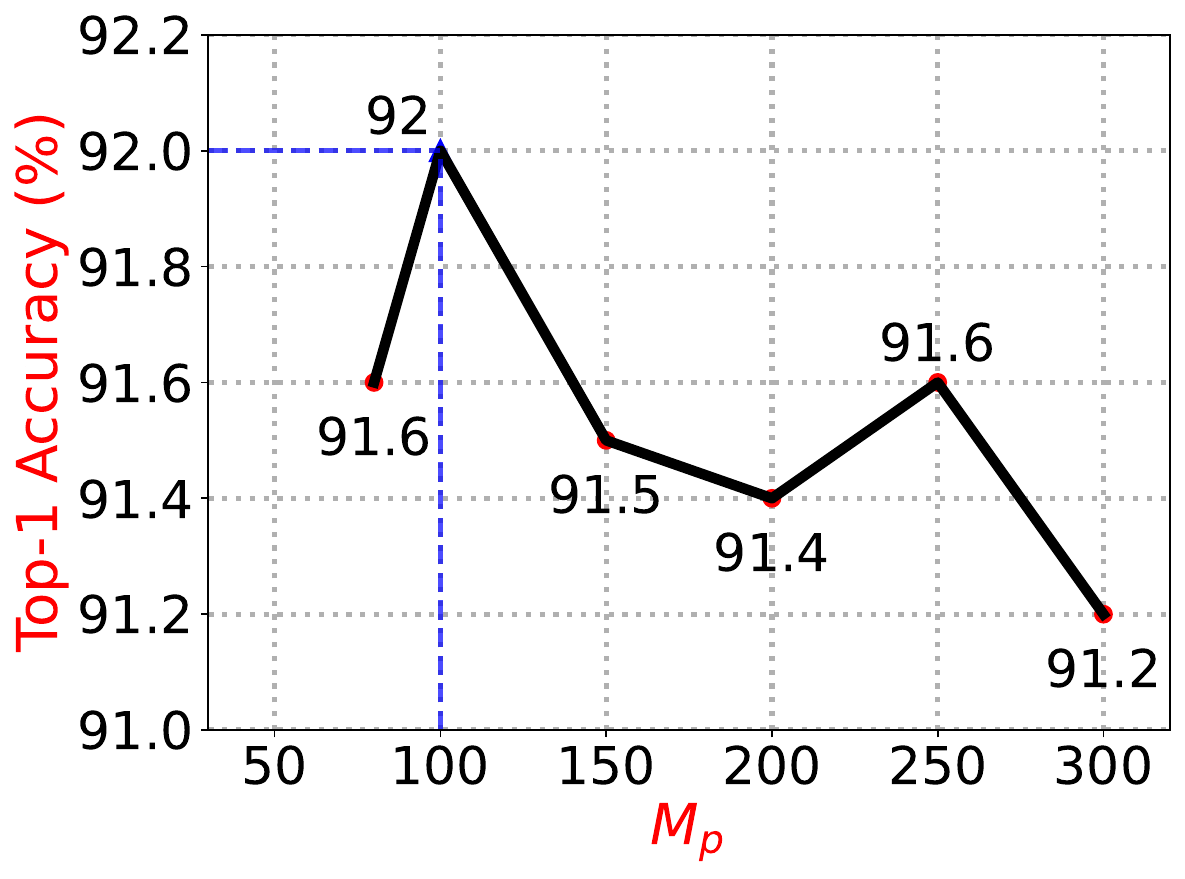}
    \end{minipage}
    \hfill
    \hspace{-7cm}
    \tikz[baseline]{\draw[dashed, thick] (0,2.8cm) -- (0,-2.8cm);}
    \hfill
    \hspace{-6cm}
     \raisebox{2\height}{%
    \begin{minipage}[t]{0.48\linewidth}
        \centering
        \includegraphics[width=0.9\linewidth]{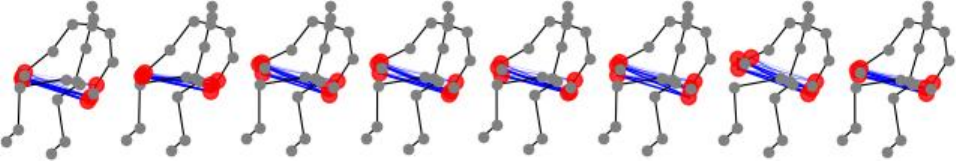}\\
        \vspace{-0.5em}
        {\small (a) Typing the keyboard}\\
        \includegraphics[width=0.9\linewidth]{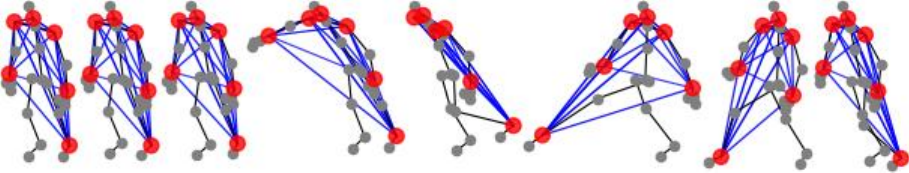}\\
        \vspace{-0.5em}
        {\small (b) Kicking something}\\
        \includegraphics[width=0.9\linewidth]{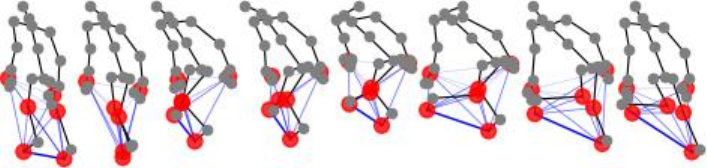}\\
        \vspace{-0.5em}
        {\small (c) Take off the shoes}\\
        \includegraphics[width=0.9\linewidth]{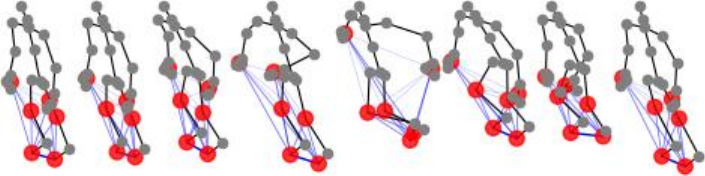}\\
        \vspace{-0.5em}
        {\small (d) Jumping up}\\
    \end{minipage}}
    \caption{Left: Ablation study on the influence of weight $\gamma$ (upper) and memory capacity $M_p$ (lower) under the NTU X-Sub setting. Right: Visualization of the learned hyperedges across 8 frames for four different action categories.}
    \label{fig:combined}
\end{figure} the complementary $\mathbf{C}_{\text{edge}}$ is essential for augmenting $\mathbf{C}_{\text{triad}}$ in capturing indirect joint relationships, and their removal leads to consistent performance degradation, confirming the necessity of both terms for high-order joint modeling.\\
\textbf{Effect of Hyperparameters.} We conduct an ablation study to further analyze the impact of different hyperparameter configurations in Contrastive Score Refinement, as illustrated in Fig.~\ref{fig:combined} (left). As a result,  we determine $\gamma=0.3$ as the optimal hyperparameter,  as it yields peak recognition accuracy. Furthermore, we examine the impact of the number of prototypes $M_p$ on recognition performance. As shown in Fig.~\ref{fig:combined} (left), accuracy tends to decrease as $M_p$ increases, which can be attributed to the fact that not all prototypes are relevant to the target action. Therefore, $M_p = 100$ is selected as the optimal configuration, as it yields the best recognition accuracy.
\vspace{-1em}
\subsection{Performance on Ambiguous Classes}
To evaluate the effectiveness of the proposed method on ambiguous action categories, we partition NTU-RGB+D into three difficulty levels based on CTR-GCN~\cite{ctrgcn} accuracy: hard ($<70\%$), medium ($70\%\text{--}90\%$), and easy ($>90\%$). The results are reported in Table~\ref{tab:underperforming}. Since ambiguous actions often fall into the hard subset, our method achieves the greatest advantage on this subset, surpassing the strongest competing method by 3.6\%, which demonstrates its effectiveness in distinguishing visually similar action classes. A more detailed analysis of each action category can be found in the supplementary material.
\subsection{Qualitative Results}
\begin{figure}[t]
    \centering
    \begin{minipage}[b]{0.48\linewidth}
        \centering
        \includegraphics[width=0.7\linewidth]{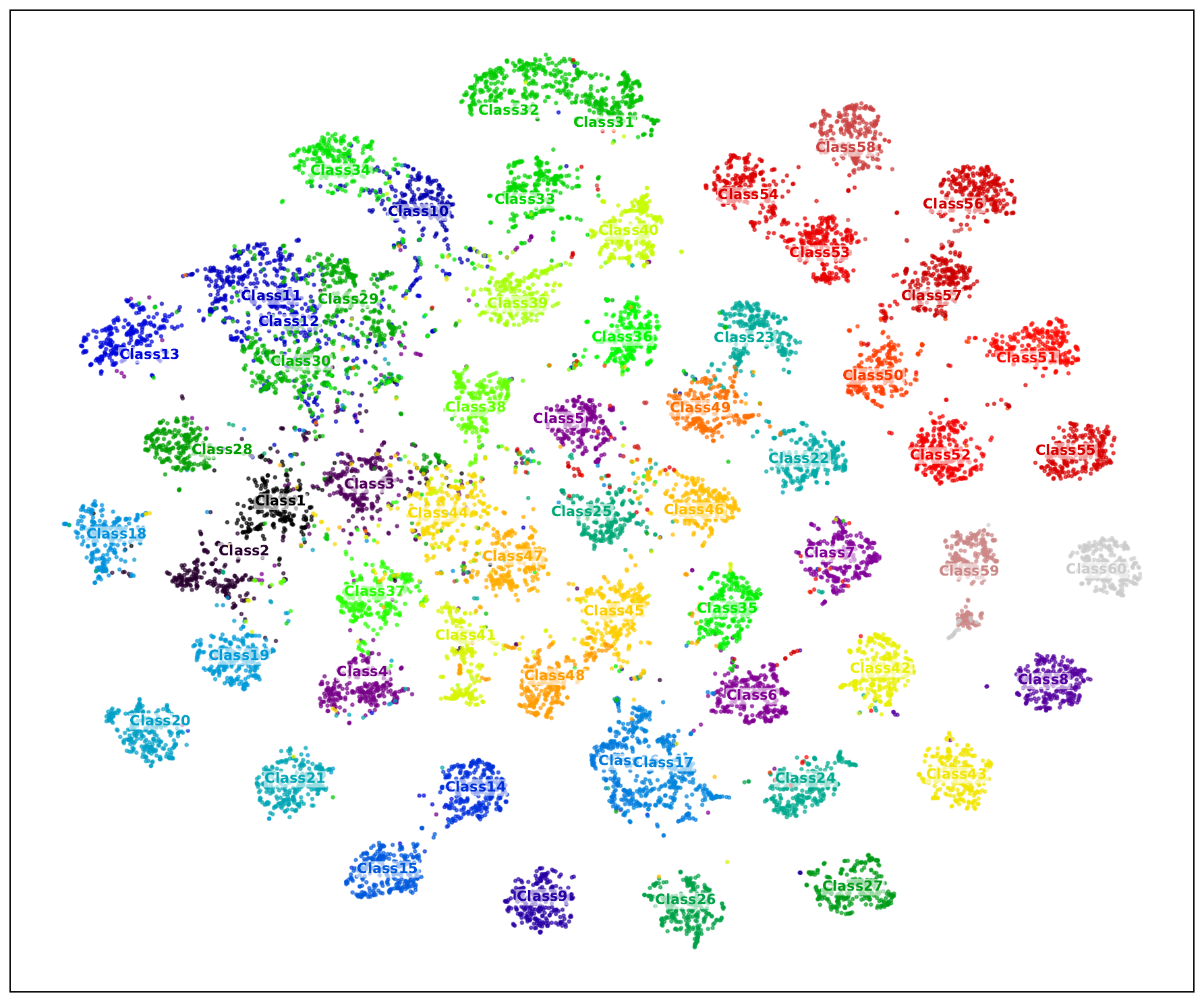}\\
        \vspace{-0.5em}
        {\small (a) Baseline}
    \end{minipage}
    \hfill
    \begin{minipage}[b]{0.48\linewidth}
        \centering
        \includegraphics[width=0.7\linewidth]{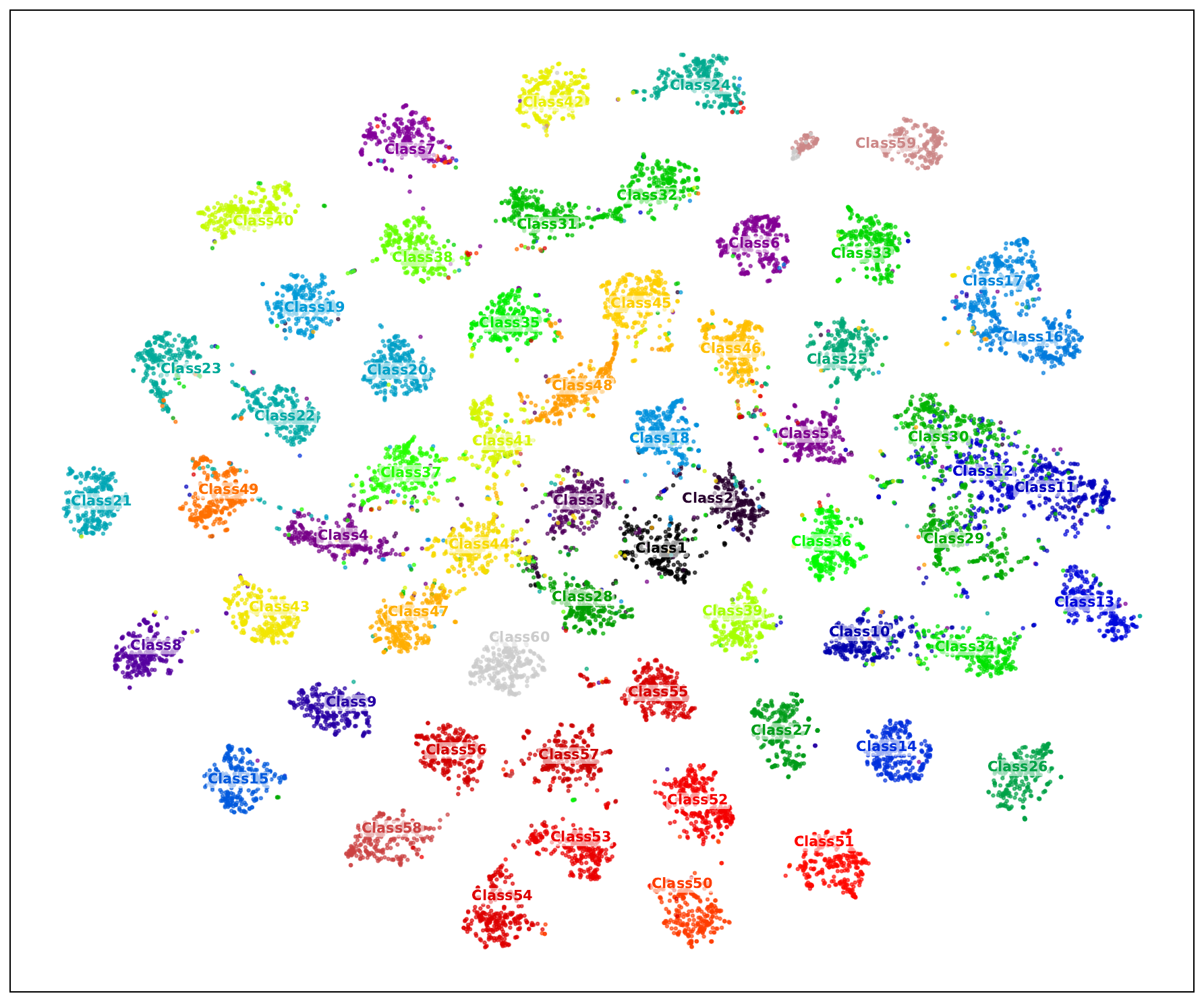}\\
        \vspace{-0.5em}
        {\small (b) DyGCN}
    \end{minipage}
    \caption{t-SNE visualization of the output features from the Baseline and DyGCN. Different colors denote different action categories.}
    \label{fig:tsne}
\end{figure}

\textbf{Visualization of the Learned Hypergraphs.} To demonstrate the effectiveness of DyGCN, we visualize the learned hyperedges for four actions: ``Typing the keyboard'', ``Kicking something'', ``Take off the shoes'', and ``Jumping up'' in Fig.~\ref{fig:combined} (right). Each action sequence is uniformly sampled every 8 frames, and each hyperedge is formed by the top-$K_1$ most compatible joints, with $K_1=5$. For ``Jumping up'' and ``Take off the shoes'', the learned hyperedges connect hands and feet, indicating coordinated interactions between non-adjacent upper- and lower-body joints. Similarly, for ``Kicking something'', the hyperedge links the right hand and left foot, which move forward simultaneously, capturing cross-limb coordination. For ``Typing the keyboard'', the learned hyperedge concentrates on the main joints involved in the action. \\
\textbf{t-SNE visualization.}
We use t-SNE to visualize the distribution in their feature space between the baseline and our method trained on the NTU RGB+D dataset, as shown in Fig.~\ref{fig:tsne}. The results indicate that DyGCN forms more compact and well-separated clusters, with clearer classification boundaries and fewer ambiguous samples compared to the baseline. This further demonstrates that the proposed method learns more discriminative action-specific representations, validating the effectiveness of both GeoHC and CSR in enhancing the separability among fine-grained actions.
\\
\section{Conclusion}
In this study, we proposed \textbf{DyGCN}, a Geometry-Constrained Dynamic Hypergraph Convolutional Network for skeleton-based action recognition. DyGCN addresses the limitations of conventional GCNs and static hypergraph methods by introducing two novel components:  the Geometry-Constrained Dynamic Hypergraph Construction (GeoHC) module and the Contrastive Score Refinement (CSR) module. Specifically, GeoHC dynamically constructs action-specific hyperedges by leveraging geometry-constrained second-order compatibility, transcending the limitations of static hyperedge structures. Furthermore, CSR refines the pairwise compatibility scores via prototype-based contrastive learning during training, fostering more discriminative action-specific hyperedges while being entirely omitted at inference. Extensive experiments on NTU RGB+D, NTU RGB+D 120, and Northwestern-UCLA demonstrate that DyGCN achieves state-of-the-art performance on widely used benchmarks.\\
Although DyGCN effectively captures dynamic high-order dependencies from standard skeleton inputs, some actions are crucially determined by semantic context, such as object interactions, action intent, and language-level motion priors. A promising future direction is to extend DyGCN with object-level context or language-guided priors from large vision-language models to further enhance dynamic hypergraph construction for ambiguous action recognition.
\section*{Acknowledgements}
This work was partly supported by the National Research Foundation of Korea (NRF) grant funded by the Korea government (MSIT) (No. 2022R1A5A8026986), Korea Institute for Advancement of Technology (KIAT) grant funded by the Korea Government (MOTIE) \\ (P0020536, HRD Program for Industrial Innovation), the Institute of Information \& Communications Technology Planning \& Evaluation (IITP)-Innovative Human Resource Development for Local Intellectualization program grant funded (80\%) by the Korea government (MSIT) (IITP-2026-RS-2020-II201462), and the Regional Innovation System \& Education (RISE) program through the Chungbuk Regional Innovation System \& Education Center (2026-RISE-11-014-03) funded by the Ministry of Education (MOE) and the
Chungcheongbuk-do (Republic of Korea).
\bibliography{egbib}
\end{document}